# Fisheye traffic data set of point center markers

Chung-I Huang[1], Wei-Yu Chen[1], Wei Jan Ko[2], Jih-Sheng Chang[1]

Chen-Kai Sun[1], Hui Hung Yu[1], Fang-Pang Lin[1]

[1]National Center for High-performance Computing

[2]National Yang Ming Chiao Tung University

Abstract

This study presents an open data-market platform and a dataset containing 160,000 markers and 18,000 images. We hope that this dataset will bring more new data value and applications In this paper, we introduce the format and usage of the dataset, and we show a demonstration of deep learning vehicle detection trained by this dataset.

I. Introduction

This study proposes an open application and challenging new traffic dataset: point-centered annotated fisheye traffic dataset. This dataset contains fisheye traffic images of four important intersections in Hsinchu City.

Traffic images of four important intersections in Hsinchu City, with the following features:

(1) the use of point-centered annotation.

(2) Three types of objects: large cars, small cars, and motorcycles.

(3) About 18,000 images and 160,000 markers are included. In addition to providing

In addition to providing the dataset, this study also develops AI models, and the model experiments prove that the dataset can provide traffic information. This dataset can provide traffic-related artificial intelligence results and applications. The dataset will be released and validated. In addition to the release and validation of the dataset, this study also proposes an open platform and model for data release.

II. Proposed Method

This study proposes an open platform and mode of data release for other needy data providers. The hope is to provide new data to bring new applications and to revitalize the data. We hope to not only provide new data to bring new applications, but also to revitalize data energy to create opportunities.

Technologies such as intelligent transportation and even autonomous driving rely heavily on a large amount of real-world data in order to develop, test, and validate AI models. Some of the traffic datasets are based on computer vision for autonomous driving, including [1-4].

Although most of the typical object detection methods are marked using the pull frame method, whether it is RCNN [5], FastRCNN [6], or Yolo [7], the cost of marking, the range of objects marked, and other conditions need to be weighed.

The current method of annotating by marking the center also has good results. CornerNet [8] detects two boundary box corners as key points, while ExtremeNet [9] detects the top, left, bottom, and right-most points, as well as the center point of all objects. CenterNet [10] simply extracts a center point for each object without grouping or post-processing.

The Fisheye traffic data set contains point center markers that indicate the location of vehicles or other objects in traffic data. These markers are typically derived from cameras, sensors, or other imaging systems, and they provide a detailed picture of the traffic behavior in a given area. The data set also includes information about the speed, direction, and other characteristics of the traffic flow. This information can be used to better understand how traffic moves in a given area and to make predictions about future traffic patterns. Recent advances in vehicle detection and tracking algorithms have enabled them to be used in a range of fields such as autonomous driving, traffic flow estimation and traffic control.

However, these models are mainly developed for rectilinear-lens cameras and are not optimized for fisheye images, which suffer from strong radial distortion, especially in the periphery. Owing to their ability to provide dynamic viewing angles and reduce blind spots, fisheye cameras are increasingly replacing traditional surveillance cameras in applications such as airports and hotels. Additionally, the field of fisheye vehicle detection and tracking is relatively new and in need of a comprehensive dataset for validating performance.

The point-centered annotated

fisheye traffic dataset is a newly proposed, open-access dataset that presents a unique challenge to researchers in the field of traffic monitoring. This dataset includes fisheye images of four intersections, each with annotated points of interest. These images provide a valuable resource for the development of cutting-edge traffic monitoring and analysis technologies.

III. Experiment

In this study, we propose a comprehensive approach to traffic data analysis and applications by releasing a dataset and developing AI models. The effectiveness of the dataset and models is demonstrated through model experiments, highlighting the potential for AI applications in traffic. To facilitate the use of this data, we also propose an open platform and model for data release. The release of a dataset and AI models for traffic data analysis allows researchers and practitioners to analyze traffic data and develop AI-based solutions for traffic-related problems. The validation of the dataset and models through experiments helps to ensure the reliability and effectiveness of the data and models for traffic data analysis.

This marker dataset is a useful tool for training and evaluating object detection models, as it includes images of three types of traffic vehicles. This dataset is helpful for developing AI applications because it allows for accurate testing of AI model performance. It contains traffic data from four intersections in Hsinchu City, including approximately 18,000 images and around 160,000 point annotations. (Fig.1)

- Use the *point center* labeling method,
- Only objects within the ROI range are marked
- The content of the label is divided into: large car, small car, locomotive

Each compressed file includes a collection of png images, json annotation files, and bound annotation images. For example:

- 01_roi.txt => mark the roi range
- frame_00000.png => original image
- frame_00000.json => annotation file
- simulation/frame_00000_out.p

ng => Binding annotated image

The marked frame_00000.png, red dots are locomotives, green dots are small cars, blue dots are large cars, the ROI range is near the intersection area within the zebra crossing, please refer to the roi.txt of each intersection for details

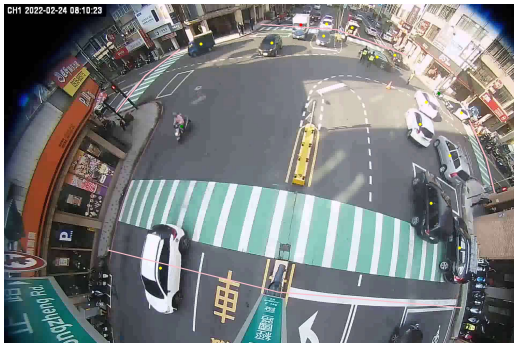

Fig.1  fisheye camera from in Hsinchu City

Each line represents a ROI line expressed as [[x1, y1], [x2, y2]], where (x1, y1) and (x2, y2) represent the two endpoints of the ROI line. (Fig.2)

```
"line": [[1732.4, 436.8], [1796.6, 582.4]]
"line": [[1796.6, 582.4], [1826.2, 708.3]]
"line": [[955.1, 76.5], [1238.9, 96.2]]
"line": [[1238.9, 96.2], [1498, 190]]
"line": [[417.1, 407.2], [525.7, 269]]
"line": [[525.7, 269], [654, 170.3]]
"line": [[387.5, 925.4], [1021.7, 1083.4]]
"line": [[1021.7, 1083.4], [1705.3, 1142.6]]
```

Fig.2 ROI line Information

```
{
  "image": {
    "file_name": "frame_01545.png",
    "width": 1920,
    "height": 1920
  },
  "points": [
    {
      "id": 264312,
      "x": 1619,
      "y": 336,
      "category_id": 0
    },
    {
      "id": 264462,
      "x": 661,
      "y": 460,
      "category_id": 0
    },
    {
      "id": 264463,
      "x": 1296,
      "y": 89,
      "category_id": 1
    },
    {
      "id": 264606,
```

Fig.3 Label data format

The annotation file is described in JSON format. The 'image' field contains the width and height and the file name. The 'points' field contains a description of the center annotation of each point. Each number has an ID, and the x,y values are the coordinates. The 'category_id' field indicates the type of transportation, where 0 is a motorcycle, 1 is a small car, and 2 is a large car. (Fig.3)

In this study, a comprehensive approach was taken to analyze the fisheye traffic dataset. The VGG19 pre-training model was utilized, along with the Adam Optimizer, Sigmoid classifier, and Loss function to compute errors. The prediction results were then obtained through visual processing. This methodology allowed for a thorough and accurate analysis of the dataset, leading to reliable and valuable insights (Fig.4).

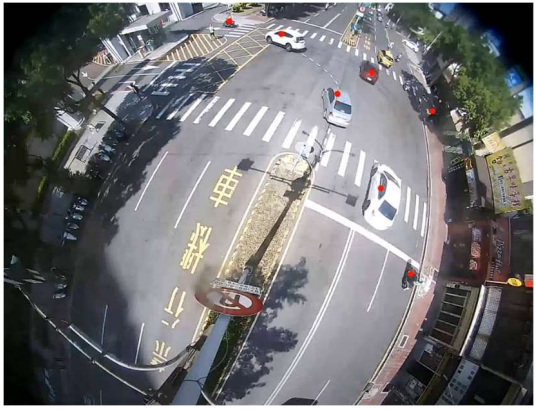

Fig.4 Fisheye data detection results (https://youtu.be/sjUQ-Ayxxtk)

IV. Data acquisition

The dataset is maintained by the National Web Center SciDM platform (http://dx.doi.org/10.30193/SciDM.DB_Fisheye_Camera_Dataset/Dataset ). iService account is required for data applicants, which is free of charge. The data owner will receive the application, review it, and notify the user that it is available for downloading, or to add additional information to the application.

V. Conclusion :

This study presents a detailed description of the fisheye traffic dataset, including its central point and the models and applications used to analyze it. Looking towards the future, we hope to continue expanding upon this dataset by releasing higher quality and larger datasets with additional features such as tracking IDs, more road segments, and longer time periods. This will provide a more comprehensive understanding of traffic patterns and behaviors, and enable the development of more effective AI-based solutions for traffic-related issues.